\documentclass[conference,a4paper]{IEEEtran}
\IEEEoverridecommandlockouts
\usepackage{cite}
\usepackage{amsmath,amssymb,amsfonts}
\usepackage{algorithmic}
\usepackage{algorithm}
\usepackage[dvips]{graphicx}
\usepackage{textcomp}
\usepackage{xcolor}
\usepackage{subfigure}
\usepackage{url}
\usepackage{listings}
\usepackage{multirow}
\newcommand{\bvec}[1]{\mbox{\boldmath $#1$}}

\def\BibTeX{{\rm B\kern-.05em{\sc i\kern-.025em b}\kern-.08em
    T\kern-.1667em\lower.7ex\hbox{E}\kern-.125emX}}

\begin{document}

\title{A Distillation Learning Model of Adaptive Structural Deep Belief Network for AffectNet: Facial Expression Image Database
\thanks{\copyright 2020 IEEE. Personal use of this material is permitted. Permission from IEEE must be obtained for all other uses, in any current or future media, including reprinting/republishing this material for advertising or promotional purposes, creating new collective works, for resale or redistribution to servers or lists, or reuse of any copyrighted component of this work in other works.}
}
\author{\IEEEauthorblockN{Takumi Ichimura}
\IEEEauthorblockA{Advanced Artificial Intelligence Project Research Center,\\
Research Organization of Regional Oriented Studies,\\
and Faculty of Management and Information System,\\
Prefectural University of Hiroshima\\
1-1-71, Ujina-Higashi, Minami-ku, \\
Hiroshima 734-8558, Japan\\
E-mail: ichimura@pu-hiroshima.ac.jp\vspace{-5mm}}
\and
\IEEEauthorblockN{Shin Kamada}
\IEEEauthorblockA{Advanced Artificial Intelligence Project Research Center,\\
Research Organization of Regional Oriented Studies,\\
Prefectural University of Hiroshima\\
1-1-71, Ujina-Higashi, Minami-ku, \\
Hiroshima 734-8558, Japan\\
E-mail: skamada@pu-hiroshima.ac.jp}
}

\maketitle

\begin{abstract}
  Deep Learning has a hierarchical network architecture to represent the complicated feature of input patterns. We have developed the adaptive structure learning method of Deep Belief Network (DBN) that can discover an optimal number of hidden neurons for given input data in a Restricted Boltzmann Machine (RBM) by neuron generation-annihilation algorithm, and can obtain the appropriate number of hidden layers in DBN. In this paper, our model is applied to a facial expression image data set, AffectNet. The system has higher classification capability than the traditional CNN. However, our model was not able to classify some test cases correctly because human emotions contain many ambiguous features or patterns leading wrong answer by two or more annotators who have different subjective judgment for a facial image. In order to represent such cases, this paper investigated a distillation learning model of Adaptive DBN. The original trained model can be seen as a parent model and some child models are trained for some mis-classified cases. For the difference between the parent model and the child one, KL divergence is monitored and then some appropriate new neurons at the parent model are generated according to KL divergence to improve classification accuracy. In this paper, the classification accuracy was improved from 78.4\% to 91.3\% by the proposed method.
\end{abstract}


\section{Introduction}
\label{sec:Introduction}
Recently, the research of Deep Learning \cite{Bengio09, Quoc12} has caused an unforeseen influence on not only theoretical research of artificial intelligence, but also practical application to realize innovations in our lives. The deep learning models such as AlexNet \cite{AlexNet}, GoogLeNet \cite{GoogLeNet15}, VGG \cite{VGG16} and ResNet \cite{ResNet} have produced image recognition tasks which greatly exceed a kind of human ability. Such models are a type of convolutional neural networks. In the models, several features included in the input data are trained at multiple layers, and such complex features of the entire image can be represented with high accuracy by piling up many layers hierarchically.

The other deep learning model is a Deep Belief Network (DBN) \cite{Hinton06} that has high classification performance by pre-training and piling up the Restricted Boltzmann Machine (RBM) \cite{Hinton12}, which is a generative stochastic neural network. Although a DBN model itself does not realize higher classification than CNN models, the automatically determined method \cite{Kamada16_SMC, Kamada16_ICONIP} of hidden neurons of RBM and hidden layers of DBN has been developed according to the input data space during learning. The method is the adaptive structural learning method of DBN (Adaptive DBN) where the neuron generation/annihilation and the layer generation algorithm work to find the optimal network structure \cite{Kamada16_TENCON}. The outstanding characteristic of Adaptive DBN can show higher classification accuracy than several CNNs \cite{Kamada18_Springer, Kamada19_WorldScientific} for some image benchmark data set \cite{CIFAR10} and the real world problems. 

In this paper, we applied the Adaptive DBN to the facial expression database AffectNet \cite{AffectNet}. AffectNet contains more than one million images with faces and extracted facial landmark points. The images are collected from the Internet and eight kinds of emotion for them are manually annotated according to the intensity of valence and arousal. The Adaptive DBN showed almost 100.0\% classification accuracy for the training data set. Although, the average classification accuracy for test data set was higher than the traditional CNN method, the accuracy for some emotion categories was not high (about 78.4\%). The mis-classification was not caused by over-fitting or over-training while learning. We found there are some conflict patterns in the test data that were categorized by two or more annotator's subjectively answers. The situation indicates when the training data is not cleansed, or when inconsistencies are found in the collected data. For such a problem, it is popular solution to cluster ambiguous data and then learn the data set by two or more models such as ensemble learning.

An ensemble learning method using multiple DBN models will be effective to realize high classification capability except the wasting computation resources because the construction of DBN will take huge iterative training. Moreover, the judgment for mis-labeled data will require other additional information.

 In this paper, we reconstruct the trained parent DBN by using the small training. First, a new child DBN is trained for the conflicted small test data. Second, Kullback-Leibler (KL) divergence is calculated to observe the difference between the parent and the child models \cite{Ichimura19}. In case of large KL, a new neuron is generated at the parent DBN and their weights are trained by the fine-tuning method described in \cite{Kamada19_WorldScientific}. We applied the proposed method to two categories which are mis-classified emotions for test data. As a result, the classification accuracy was improved from 78.4\% to 91.3\%.

The remainder of this paper is organized as follows. In section II, the adaptive structural learning of DBN is briefly explained. The section III explains the facial database: AffectNet. In section IV, the effectiveness of our proposed method is verified on AffectNet. In section V, we give some discussions to conclude this paper.

\section{Adaptive Structural Learning Method of DBN}
\label{sec:Adaptive_dbn}
\begin{figure}[]
\centering
\includegraphics[scale=0.4]{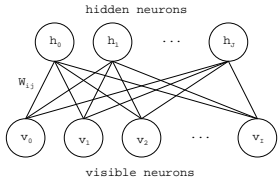}
\vspace{-3mm}
\caption{Network structure of RBM}
\label{fig:rbm}
\vspace{-5mm}
\end{figure}

The basic idea of our proposed Adaptive DBN is described in this section to understand the basic behavior of self-organized structure briefly.
A RBM \cite{Hinton12} is a stochastic unsupervised learning model. The network structure of RBM consists of two kinds of binary layers as shown in Fig.~\ref{fig:rbm}. DBN is a stacking box for stochastic unsupervised learning by hierarchically building several pre-trained RBMs. 

\subsection{Neuron Generation/Annihilation Algorithm}
\label{subsec:adaptive_rbm}
\begin{figure}[]
\begin{center}
\subfigure[Neuron generation]{\includegraphics[scale=0.4]{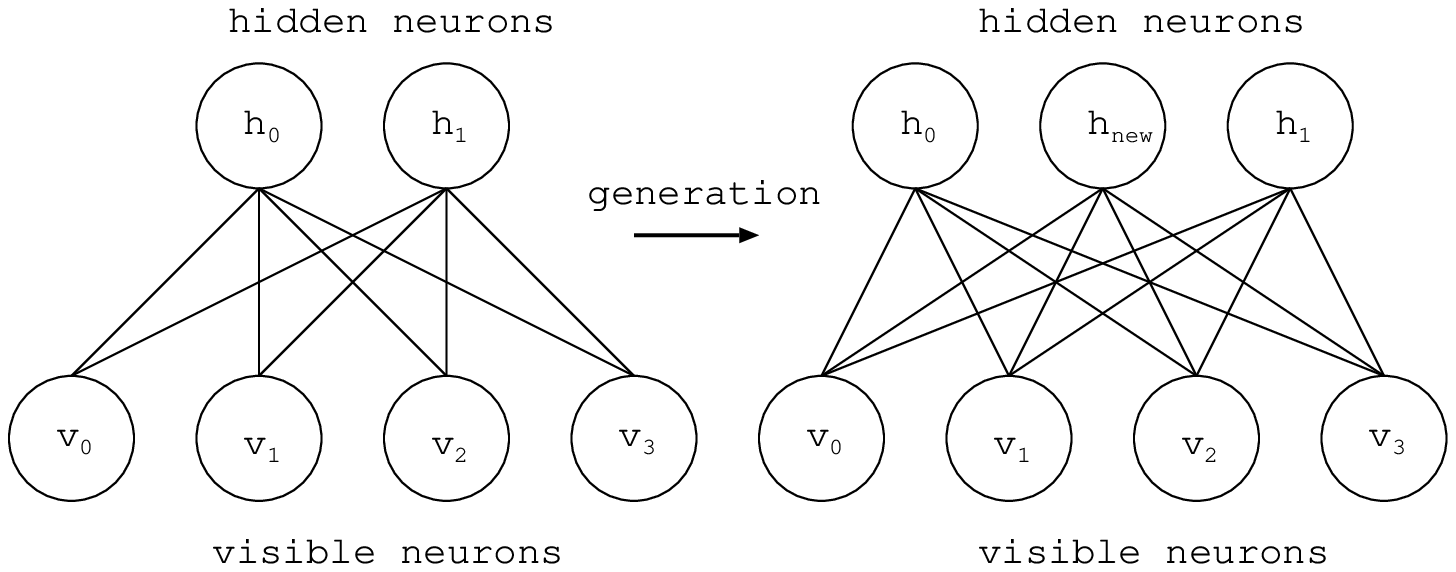}\label{fig:neuron_generation}}
\subfigure[Neuron annihilation]{\includegraphics[scale=0.4]{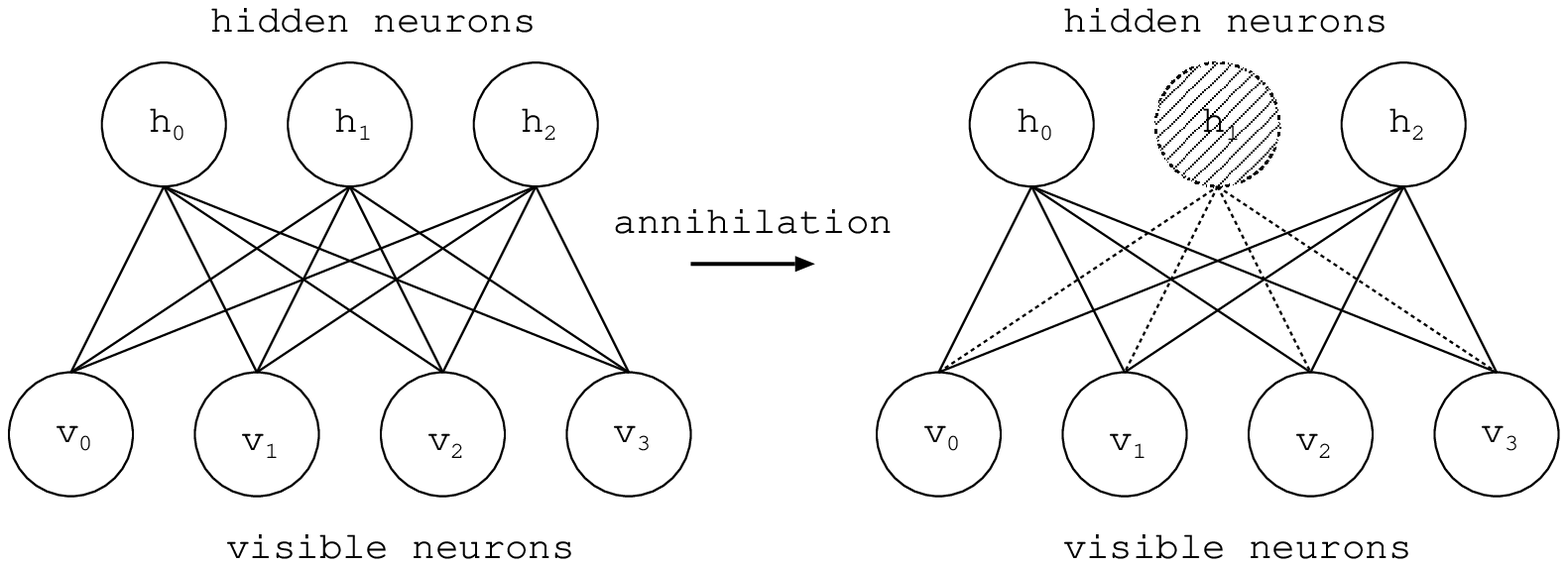}\label{fig:neuron_annihilation}}
\vspace{-3mm}
\caption{Adaptive RBM}
\label{fig:adaptive_rbm}
\vspace{-5mm}
\end{center}
\end{figure}

We have developed the adaptive structural learning method in RBM and DBN, called Adaptive RBM and Adaptive DBN \cite{Kamada18_Springer}. The neuron generation algorithm of Adaptive RBM is able to generate an optimal number of hidden neurons for given input space during its training situation. In the method, the idea of Walking Distance (WD) was proposed as the difference between the past variance and the current variance for learning parameters. If the network does not have enough neurons to classify them sufficiently, then the WD will tend to fluctuate largely after the long training process. The situation shows that some hidden neurons may not represent an ambiguous pattern due to the lack of the number of hidden neurons. In order to represent ambiguous patterns into two neurons, a new neuron is inserted to inherit the attributes of the parent hidden neuron as shown in Fig.~\ref{fig:neuron_generation}.

However, we may meet a situation that some unnecessary or redundant neurons were generated due to the neuron generation process. The neuron annihilation algorithm was applied to kill the corresponding neuron after neuron generation process. Fig.~\ref{fig:neuron_annihilation} shows that the corresponding neuron is annihilated.

\begin{figure}[]
\centering
\includegraphics[scale=0.5]{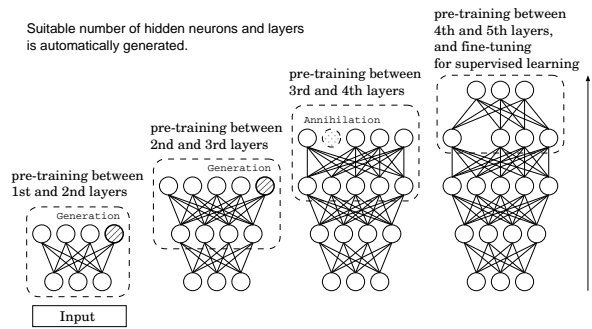}
\vspace{-3mm}
\caption{An Overview of Adaptive DBN}
\label{fig:adaptive_dbn}
\vspace{-5mm}
\end{figure}

DBN is a hierarchical model of stacking the several pre-trained RBMs. For building hierarchical process, output (activation of hidden neurons) of $l$-th RBM can be seen as the next input of $l+1$-th RBM. Generally, DBN with many RBMs has higher power of data representation than one RBM. Such hierarchical model can represent the various features from an abstract concept to concrete representation at each layer in the direction of input layer to output layer. However, the optimal number of RBMs depends on the target data space.

Adaptive DBN can automatically adjust an optimal network structure to add a new RBM layer one by one based on the idea of WD. If both WD and the energy function do not become small values, a new RBM will be generated to make the network structure suitable for the data set, since the RBM has lacked data representation capability to figure out an image of input patterns. Therefore, the condition for layer generation is defined by using the total WD at the $l$-th layer, $WD^{l}$, and the energy function, $E^{l}$. Fig.~\ref{fig:adaptive_dbn} shows the overview of layer generation in Adaptive DBN (For details see \cite{Kamada18_Springer}).

\section{Adaptive DBN for AffectNet}
\label{sec:AffectNet}
\begin{figure}[tbp]
  \centering
  \subfigure[Neutral]{\includegraphics[width=16mm]{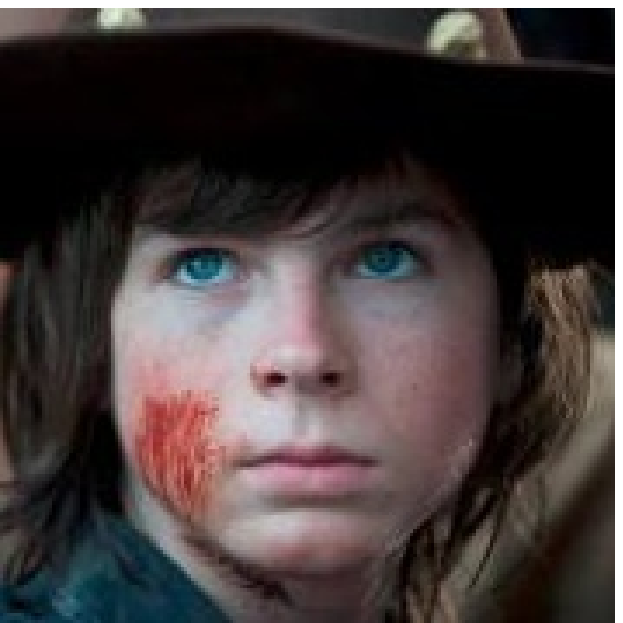}\label{fig:sample_0}}
  \subfigure[Happy]{\includegraphics[width=16mm]{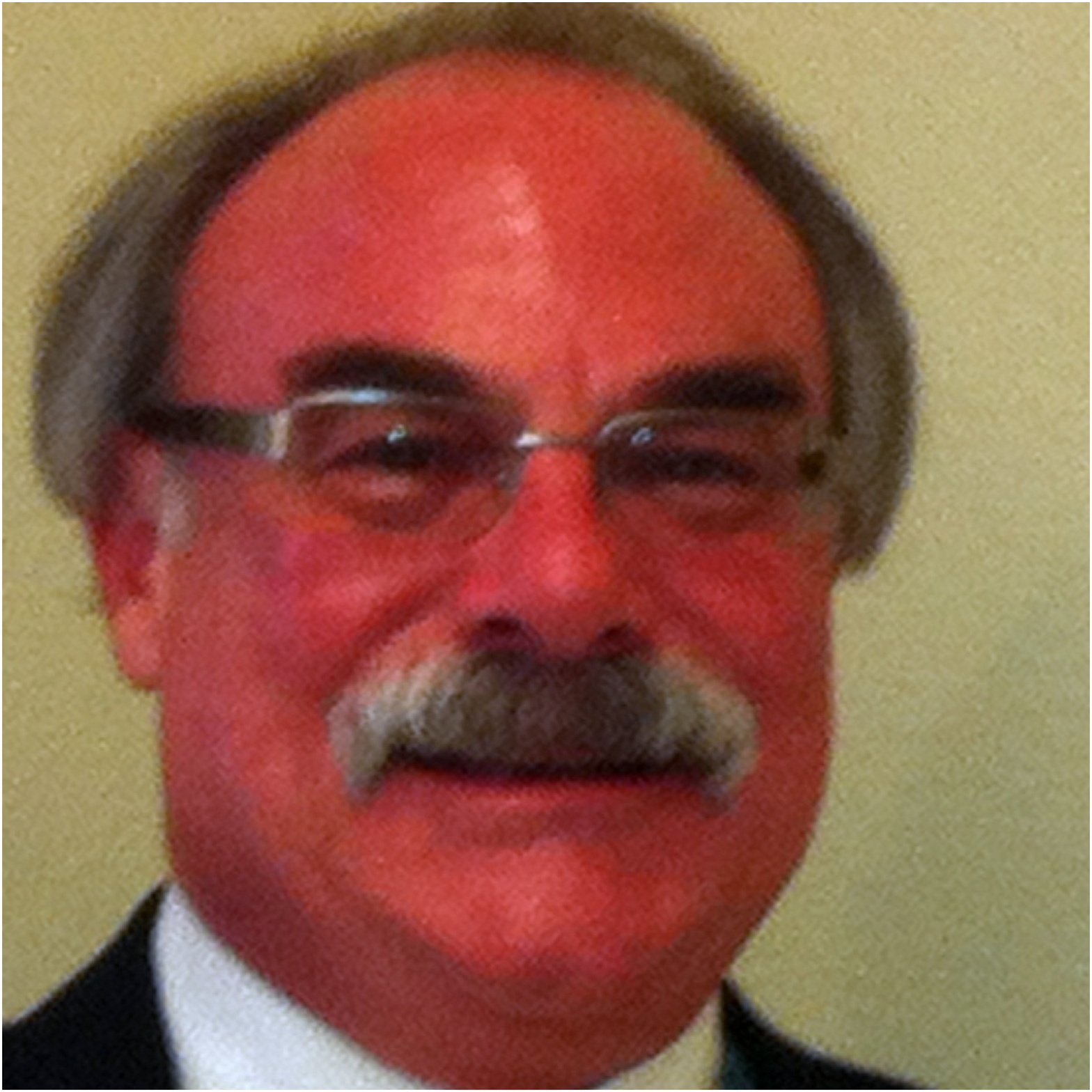}\label{fig:sample_1}}
  \subfigure[Sad]{\includegraphics[width=16mm]{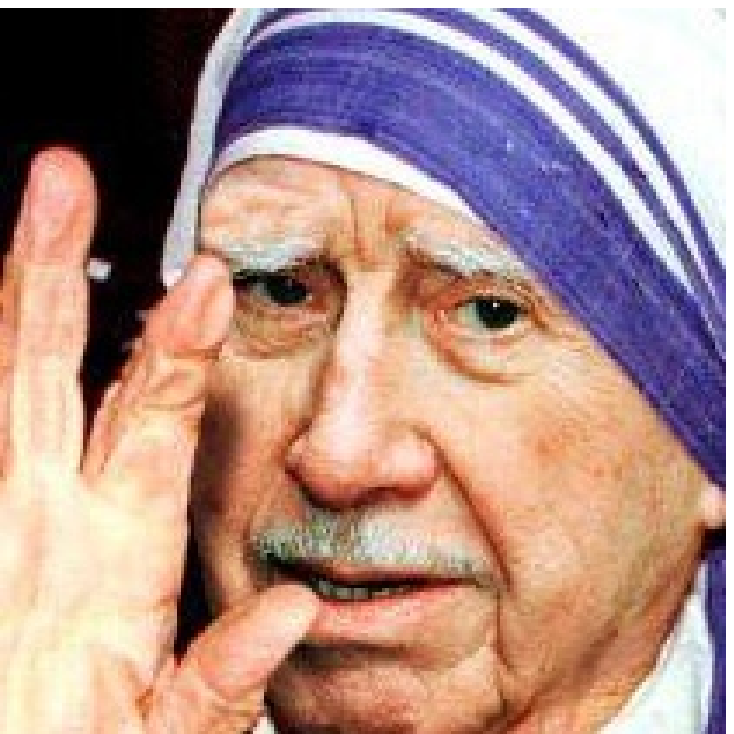}\label{fig:sample_2}}
  \subfigure[Surprise]{\includegraphics[width=16mm]{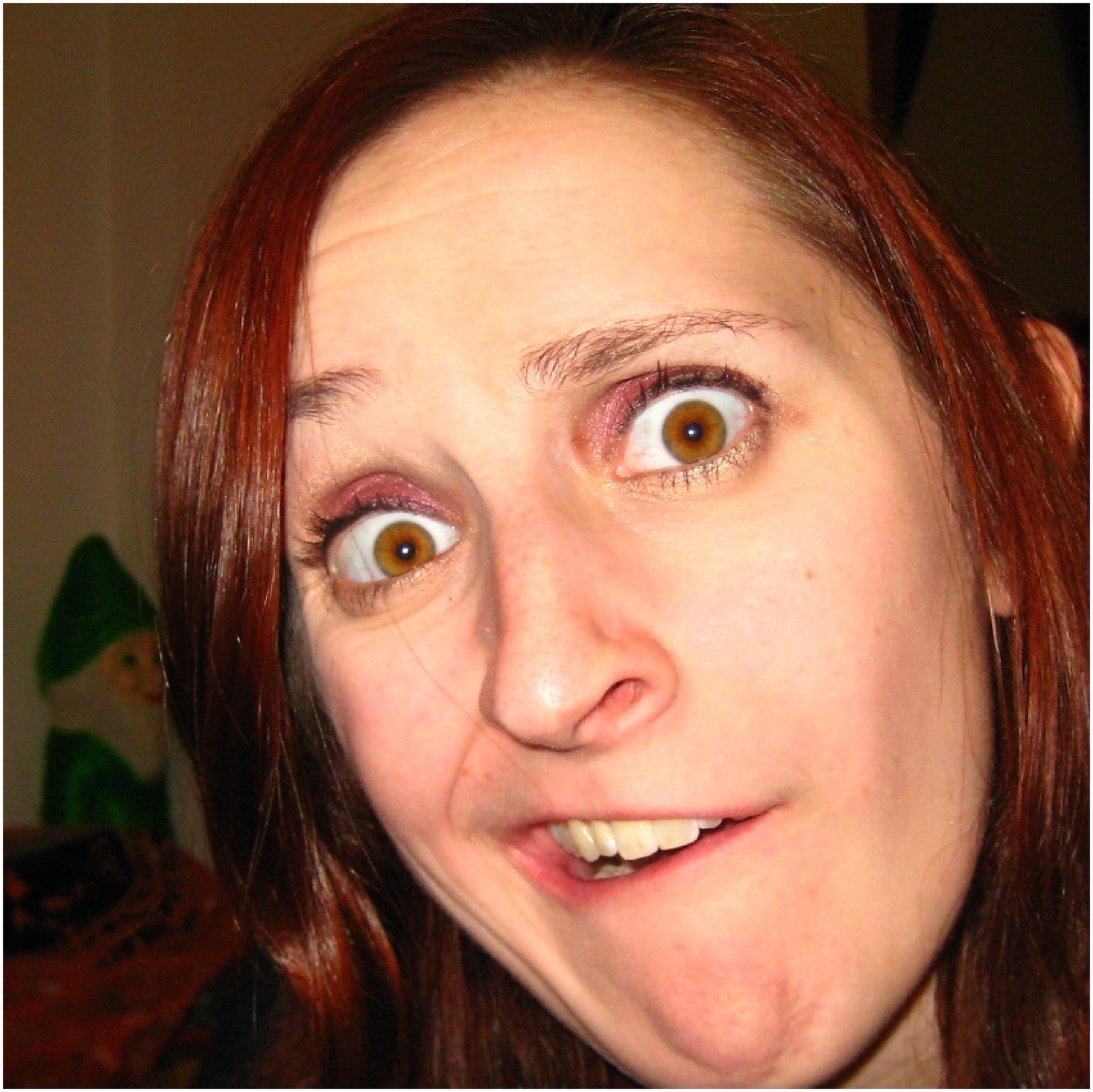}\label{fig:sample_3}}
  \subfigure[Fear]{\includegraphics[width=16mm]{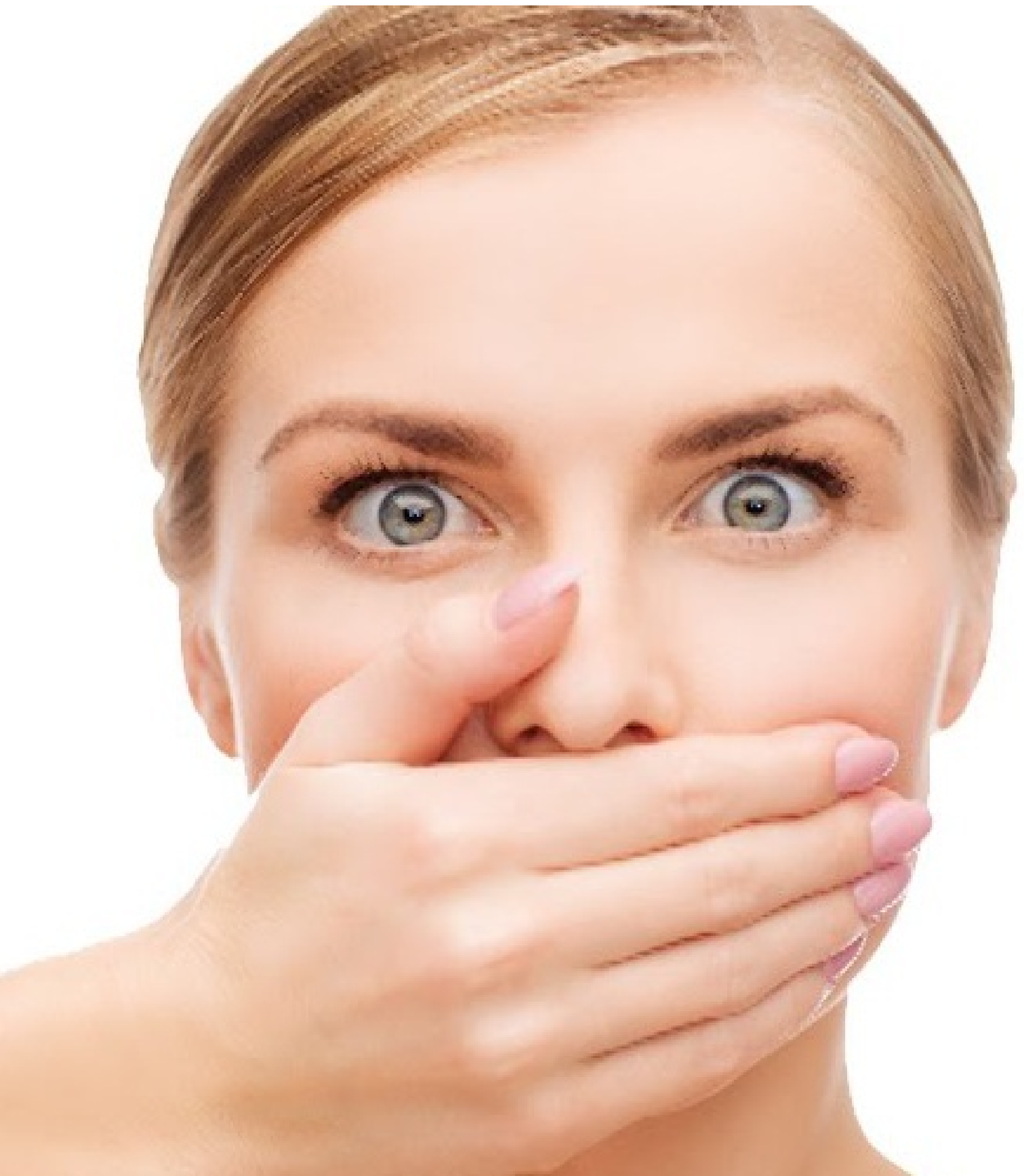}\label{fig:sample_4}}
  \subfigure[Disgust]{\includegraphics[width=16mm]{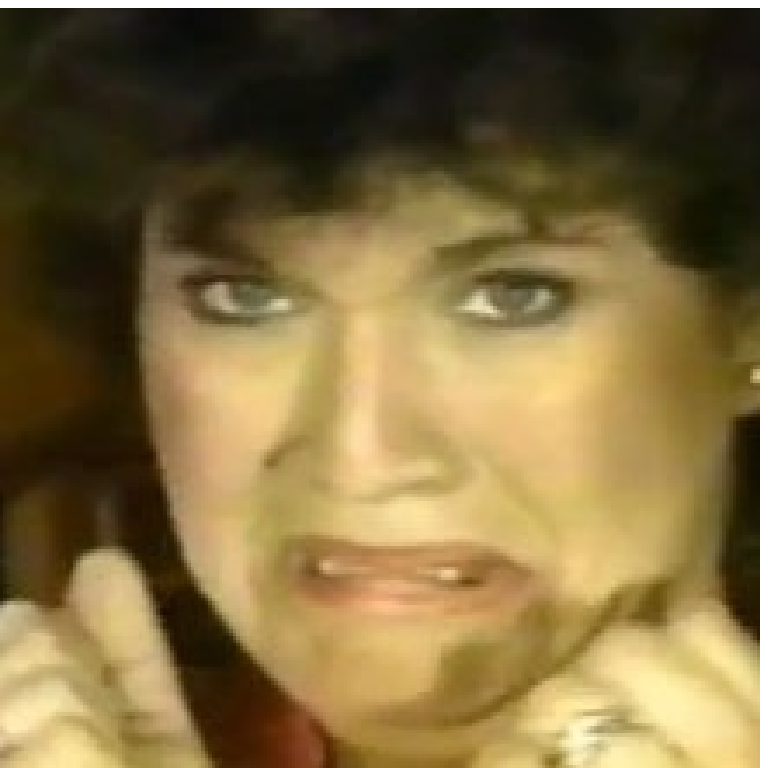}\label{fig:sample_5}}
  \subfigure[Anger]{\includegraphics[width=16mm]{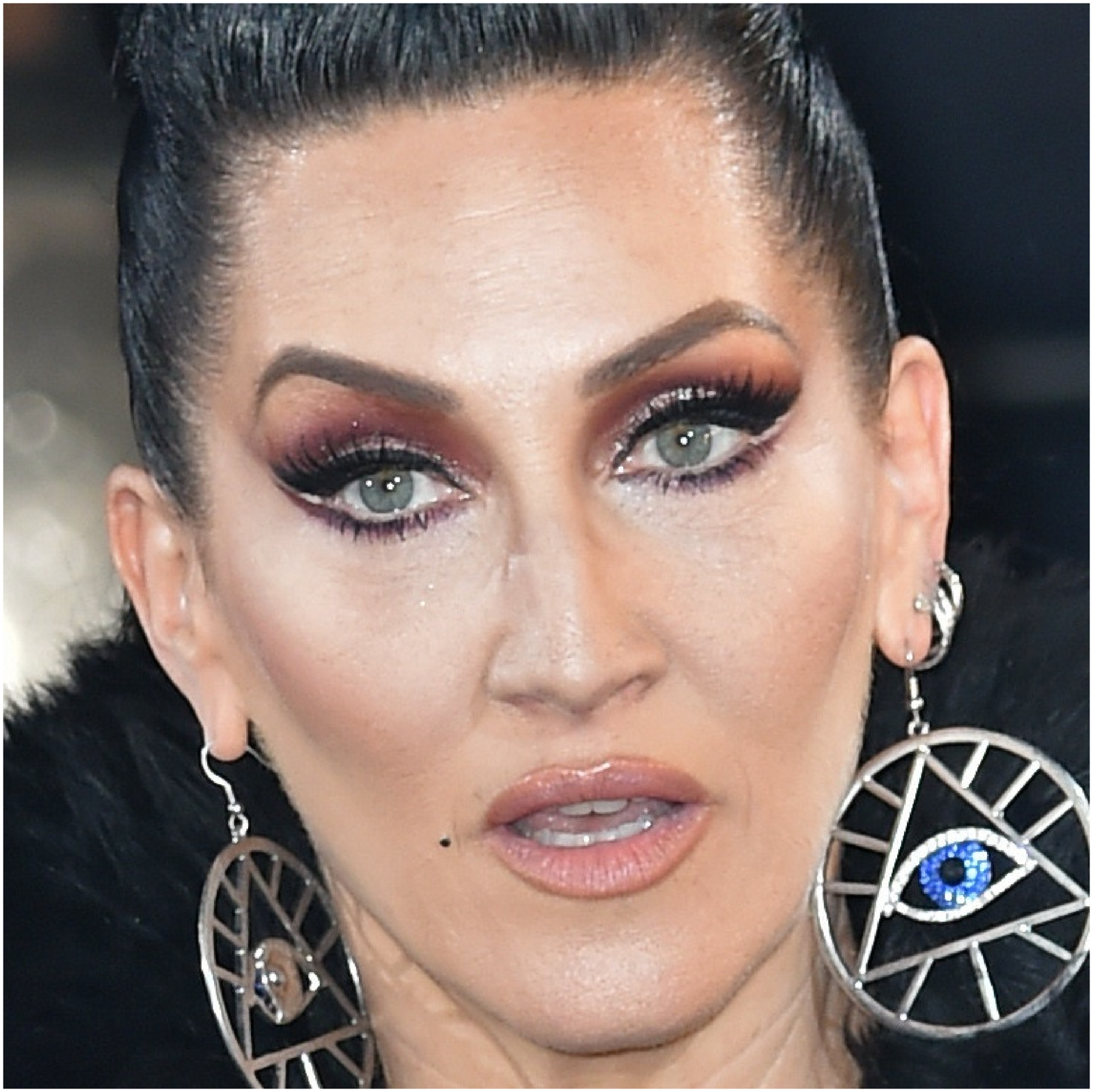}\label{fig:sample_6}}
  \subfigure[Contempt]{\includegraphics[width=16mm]{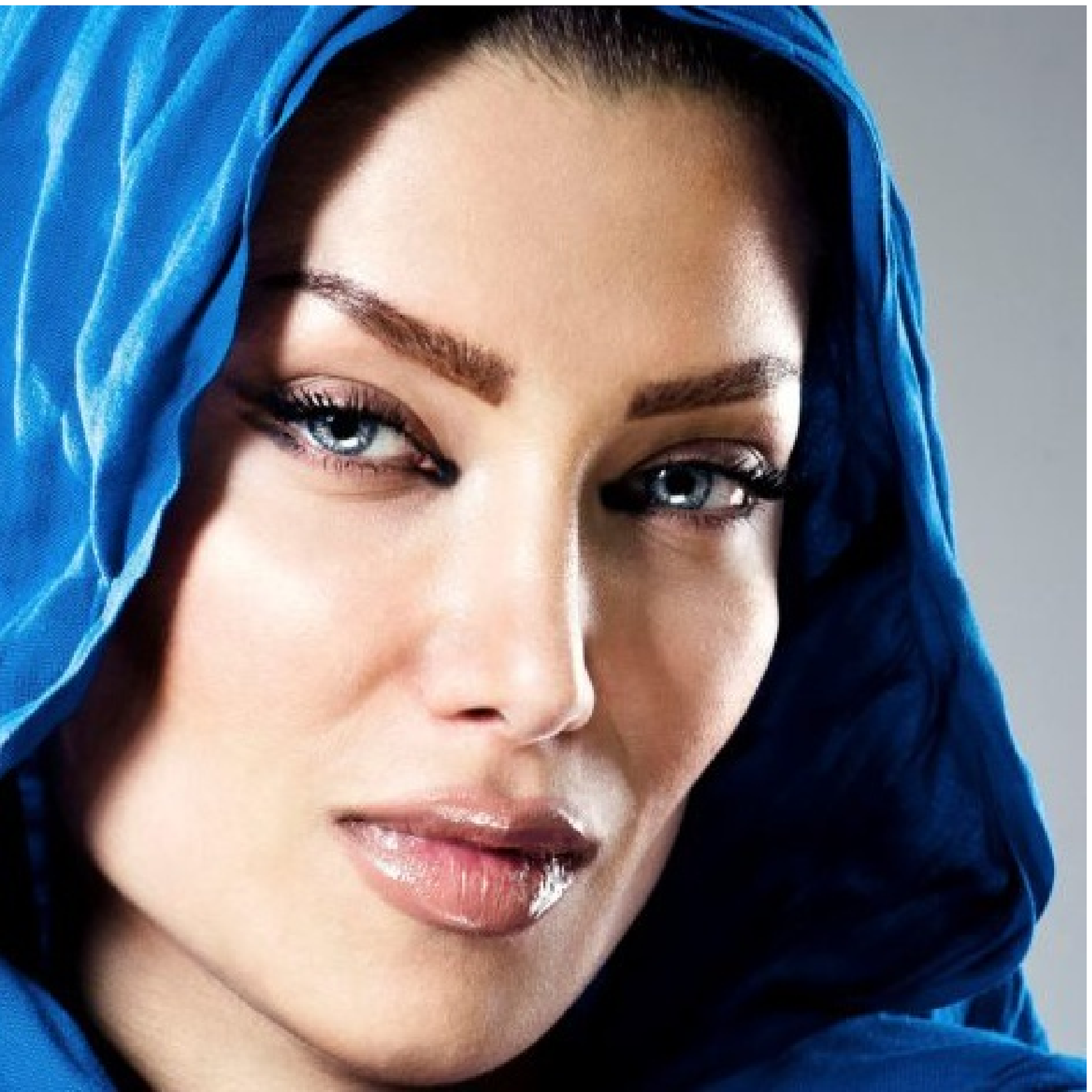}\label{fig:sample_7}}
  \subfigure[None]{\includegraphics[width=16mm]{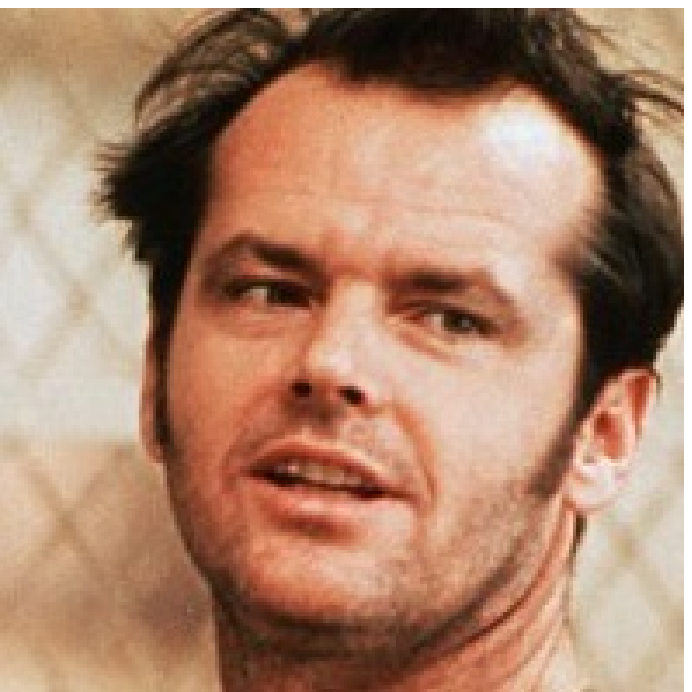}\label{fig:sample_8}}
  \subfigure[Uncertain]{\includegraphics[width=16mm]{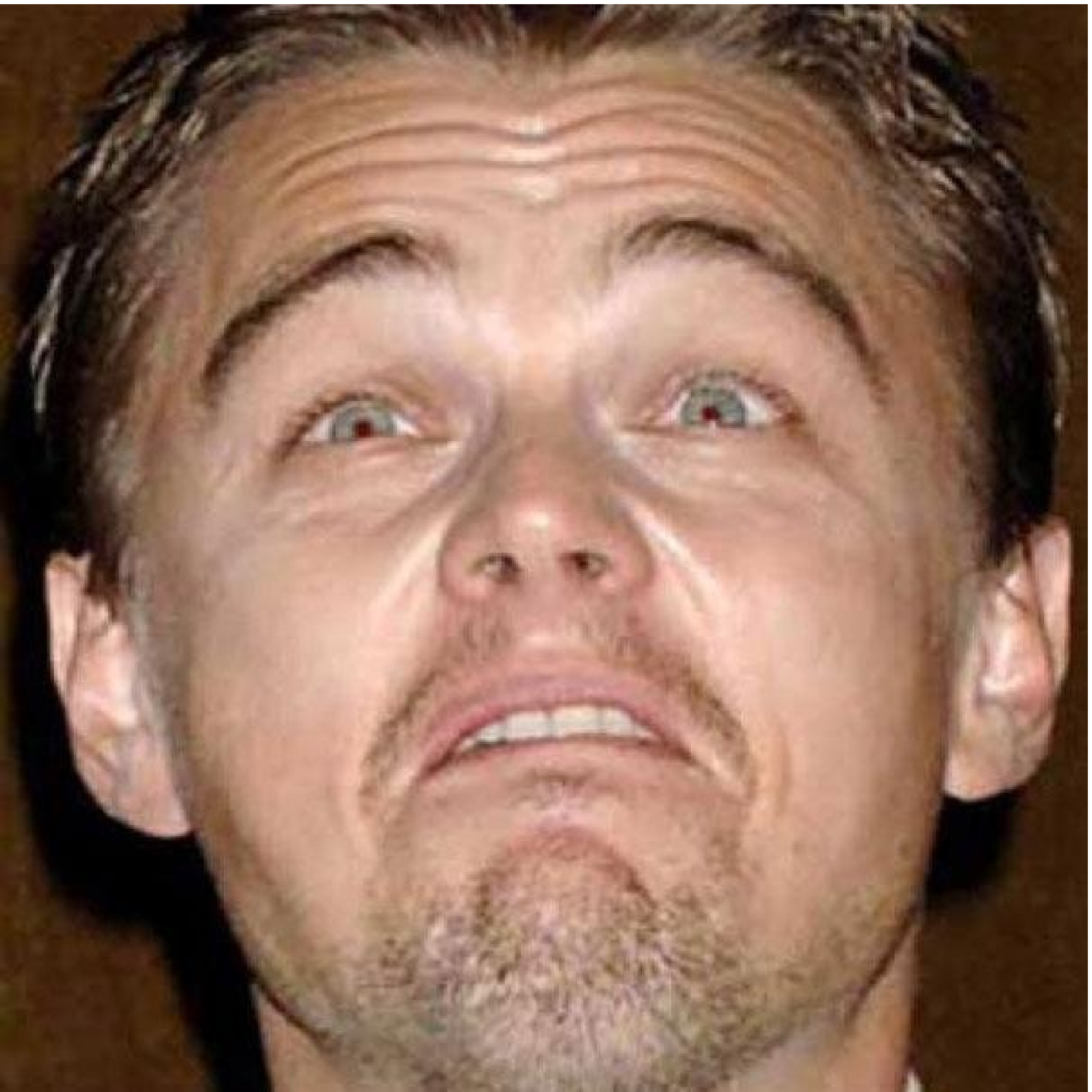}\label{fig:sample_9}}
  \vspace{-3mm}
  \caption{Samples in AffectNet}
  \label{fig:data_affectnet_sample}
\vspace{-5mm}
\end{figure}

\begin{figure}[tbp]
  \centering
  \includegraphics[scale=0.24]{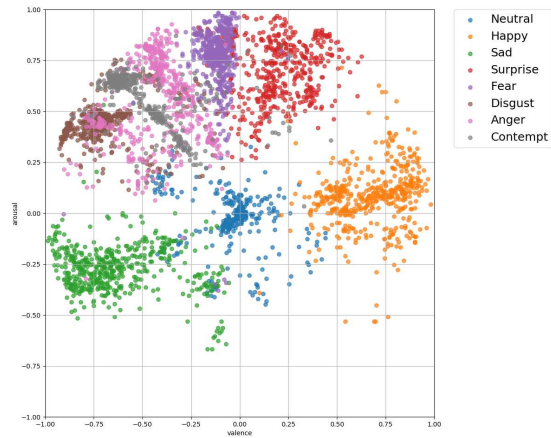}
\vspace{-3mm}
  \caption{Valence and Arousal}
  \label{fig:valence_arousal}
\vspace{-3mm}
\end{figure}

AffectNet is a facial image database containing human emotions collected from the Internet\cite{AffectNet}. As shown in Table~\ref{tab:data_affectnet_category}, there are eleven categories including training cases and test cases. Fig.~\ref{fig:data_affectnet_sample} shows the sample images for ten categories except `Non-Face'. These categories are labeled according to human subjectivity based on Valence and Arousal as shown in Fig.~\ref{fig:valence_arousal}. Valence is the degree of pleasure/displeasure and Arousal is the degree of emotional strength. The value of Valence and Arousal takes in $[-1,1]$, respectively as shown in Fig.~\ref{fig:valence_arousal}. Fig.~\ref{fig:valence_arousal} shows the distribution of Valence and Arousal for eight emotion categories.

An Adaptive DBN was trained for training cases of eight categories except the three categories `None,' `Uncertain,' and `Non-Face' in the same way of \cite{AffectNet}. Although, the average classification accuracy for test data set was higher than the traditional CNN method, the accuracy for some emotion categories was not high (about 78.4\%). The mis-classification was not caused by over-fitting or over-training while learning. We found there are some conflict patterns in the test data that were categorized by two or more annotator's subjective answers.

According to \cite{AffectNet}, two annotators labeled the category for same facial images, but their annotation results are not same. Table~\ref{tab:ref} shows the agreement matrix between two categories labeled by the two annotators. Table~\ref{tab:ref} seems to be a natural result for the decision process of emotion types to many facial images without evaluating or adjusting criterion among annotators. Since human emotions contain many vague features, each annotator's answer can not be always same. For example, a decision may include an annotator's preference to the celebrities images in the labeling process. In order to classify such data automatically, an ensemble mechanism of deep learning will be required. An ensemble learning method using multiple DBN models will be effective to realize high classification capability except the wasting computation resources because the construction of DBN will take huge iterative training. Moreover, the judgment for mis-labeled data will require other additional information. Two or more DBNs were trained to improve the classification ratio in the following section.

\begin{table}[tb]
\caption{Emotional Category in AffectNet}
\vspace{-3mm}
\label{tab:data_affectnet_category}
\begin{center}
\scalebox{0.8}[0.8]{
\begin{tabular}{l|r|r}
\hline \hline
\multicolumn{1}{c|}{Category} & Train Data & Test Data \\ \hline
\hline
Neutral   &  74,874  & 500 \\ \hline
Happy     & 134,415 &  500 \\ \hline
Sad       &  25,459 &  500 \\ \hline
Surprise  &  14,090 &  500 \\ \hline
Fear      &   6,378 &  500 \\ \hline
Disgust   &   3,803 &  500 \\ \hline
Anger     &  24,882 &  500 \\ \hline
Contempt  &   3,750 &  500 \\ \hline
None      &  33,088 &  500 \\ \hline
Uncertain &  11,645 &  500 \\ \hline
Non-Face  &  82,414 &  500 \\ \hline\hline
\multicolumn{1}{c|}{Total} & 414,798 & 5500\\ \hline
\hline
\end{tabular}
} 
\end{center}
\vspace{-5mm}
\end{table}

\begin{table*}[tbp]
\caption{The Category of Agreement Between Two Annotators in Categorical Model of Affect (\%) \cite{AffectNet}}
\vspace{-3mm}
\label{tab:ref}
\begin{center}
\scalebox{0.8}[0.8]{
\begin{tabular}{l|r|r|r|r|r|r|r|r|r|r|r}
\hline \hline
& Neutral & Happy & Sad & Surprise & Fear & Disgust & Anger & Contempt & None & Uncertain & Non-Face\\ \hline \hline
Neutral & 50.8 & 7 & 9.1 & 2.8 & 1.1 & 1 & 4.8 & 5.3 & 11.1 & 1.9 & 5.1 \\ \hline
Happy & 6.3 & 79.6 & 0.6 & 1.7 & 0.3 & 0.4 & 0.5 & 3 & 4.6 & 1 & 2.2 \\ \hline
Sad & 11.8 & 0.9 & 69.7 & 1.2 & 3.4 & 1.3 & 4 & 0.3 & 3.5 & 1.2 & 2.6 \\ \hline
Surprise & 2 & 3.8 & 1.6 & 66.5 & 14 & 0.8 & 1.9 & 0.6 & 4.2 & 1.9 & 2.7 \\ \hline
Fear & 3.1 & 1.5 & 3.8 & 15.3 & 61.1 & 2.5 & 7.2 & 0 & 1.9 & 0.4 & 3.3 \\ \hline
Disgust & 1.5 & 0.8 & 3.6 & 1.2 & 3.5 & 67.6 & 13.1 & 1.7 & 2.7 & 2.3 & 2.1 \\ \hline
Anger & 8.1 & 1.2 & 7.5 & 1.7 & 2.9 & 4.4 & 62.3 & 1.3 & 5.5 & 1.9 & 3.3 \\ \hline
Contempt & 10.2 & 7.5 & 2.1 & 0.5 & 0.5 & 4.4 & 2.1 & 66.9 & 3.7 & 1.5 & 0.6 \\ \hline
None & 22.6 & 12 & 14.5 & 8 & 6 & 2.3 & 16.9 & 1.3 & 9.6 & 4.3 & 2.6 \\ \hline
Uncertain & 13.5 & 12.1 & 7.8 & 7.3 & 4 & 4.5 & 6.2 & 2.6 & 12.3 & 20.6 & 8.9 \\ \hline
Non-Face & 3.7 & 3.8 & 1.7 & 1.1 & 0.9 & 0.4 & 1.7 & 0.4 & 1.2 & 1.4 & 83.9 \\ \hline
\hline
\end{tabular}
}
\end{center}
\vspace{-5mm}
\end{table*}

\section{KL divergence based Distillation model}
\label{sec:KLDivergence_EnsembleDBN}
The section describes the distillation learning model by using KL Divergence of Adaptive DBN. The classification accuracy for test cases was investigated in each category. The mis-classification for some specified emotional types was occurred by two annotators' subjectivity described in Section \ref{sec:AffectNet}. The KL divergence between the trained model and the mis-classified emotional type was calculated. If the KL divergence is larger than the certain threshold value, some generated new neurons will be required at the appropriate position to classify the mis-classified cases. The neuron generation method was implemented to patch the wrong path of the network.

\subsection{Classification accuracy of AffectNet by Adaptive DBN}
\label{sec:Classification_AffectNet}
Table \ref{tab:classification_ratio} shows the classification results by Adaptive DBN for AffectNet. In \cite{AffectNet}, the classification result by AlexNet was reported about 60\% of the test data set. Adaptive DBN showed about 100.0\% classification accuracy for the training data set. For test data set, the classification accuracy for eight emotion types was 87.4\% on average. The best category was `Happy', while the worst one was `Anger'. The categories with less than 90\% classification accuracy were `Neutral,' `Sad,' `Surprise,' `Anger,' and `Contempt.' We investigated the relation between the label and the facial images for the mis-classified categories.

\begin{figure}[tbp]
  \centering
  \subfigure[Disgust $\rightarrow$ Anger]{\includegraphics[width=13mm]{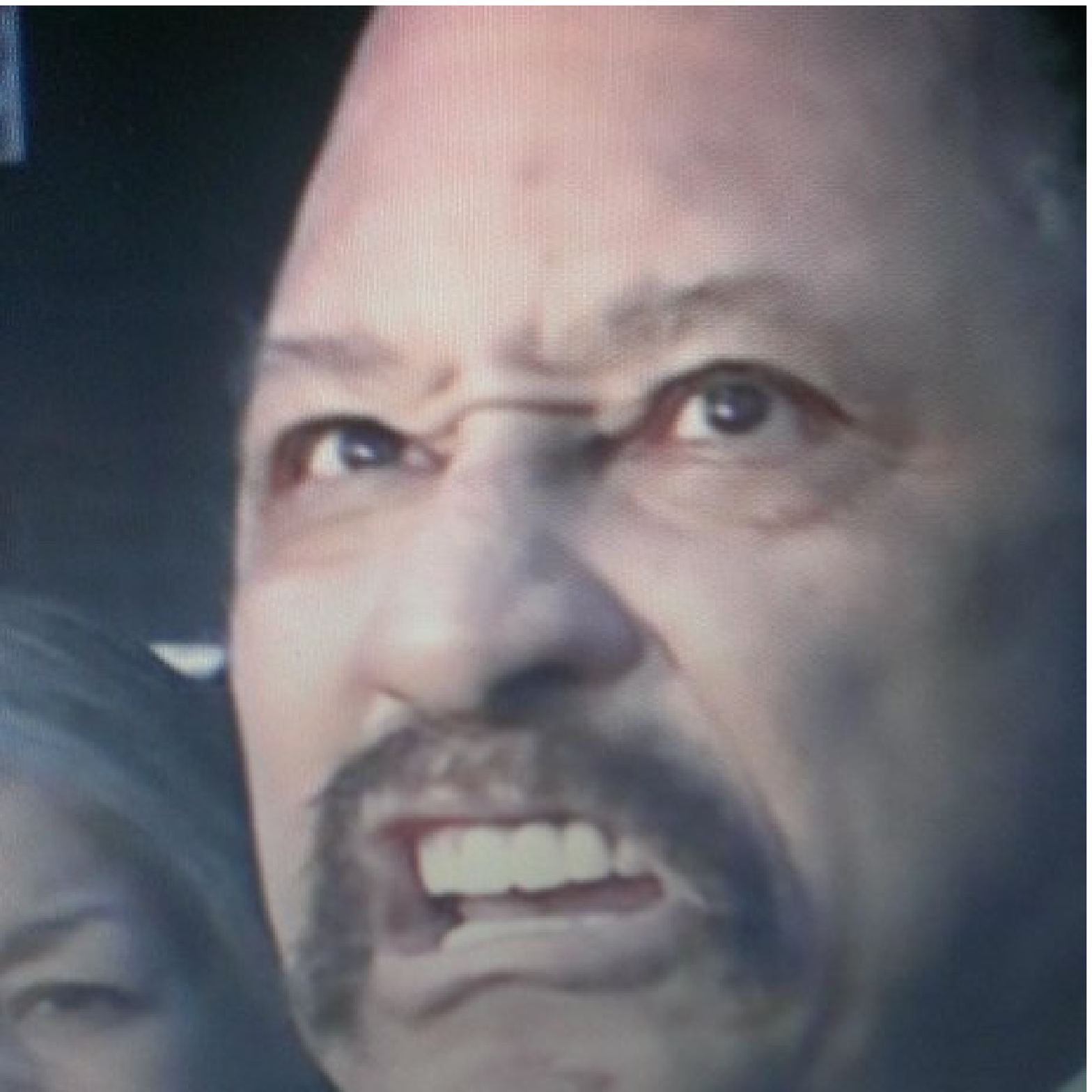}\label{fig:WA1}}
  \subfigure[Disgust $\rightarrow$ Anger]{\includegraphics[width=13mm]{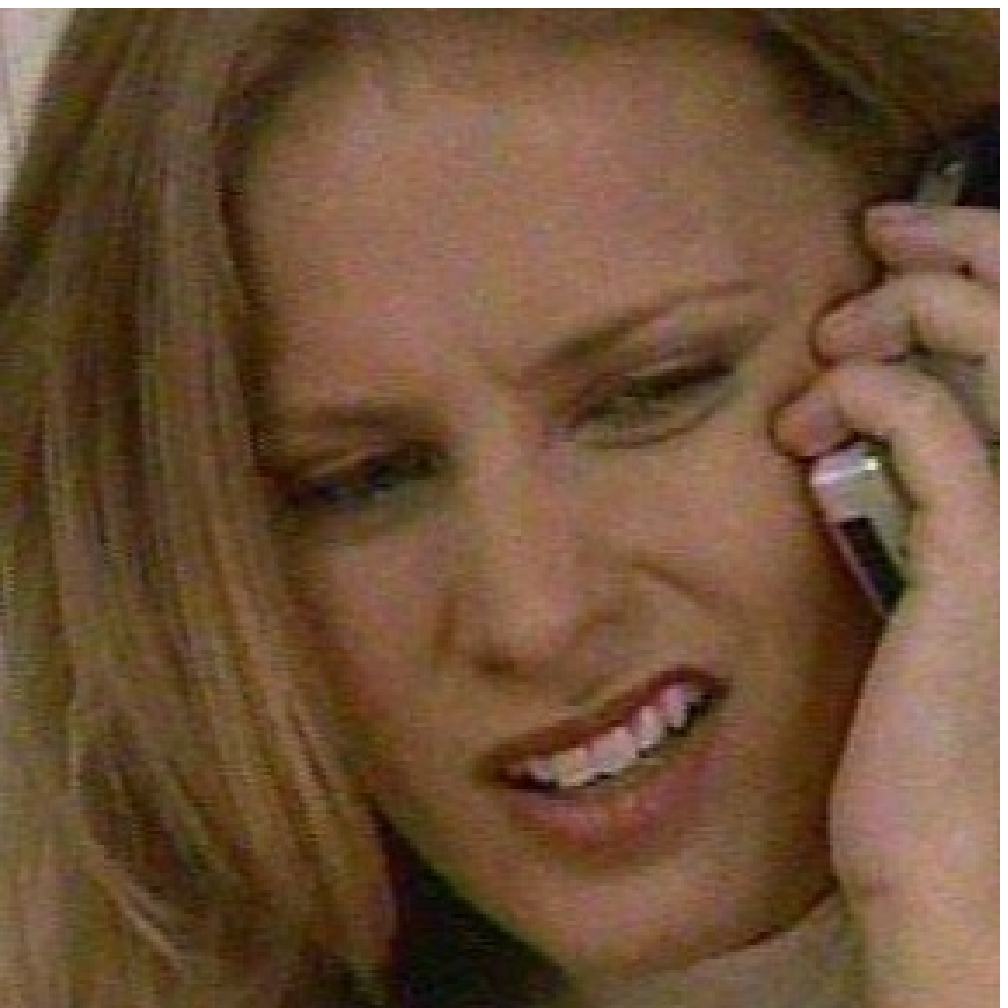}\label{fig:WA2}}
  \subfigure[Disgust $\rightarrow$ Anger]{\includegraphics[width=13mm]{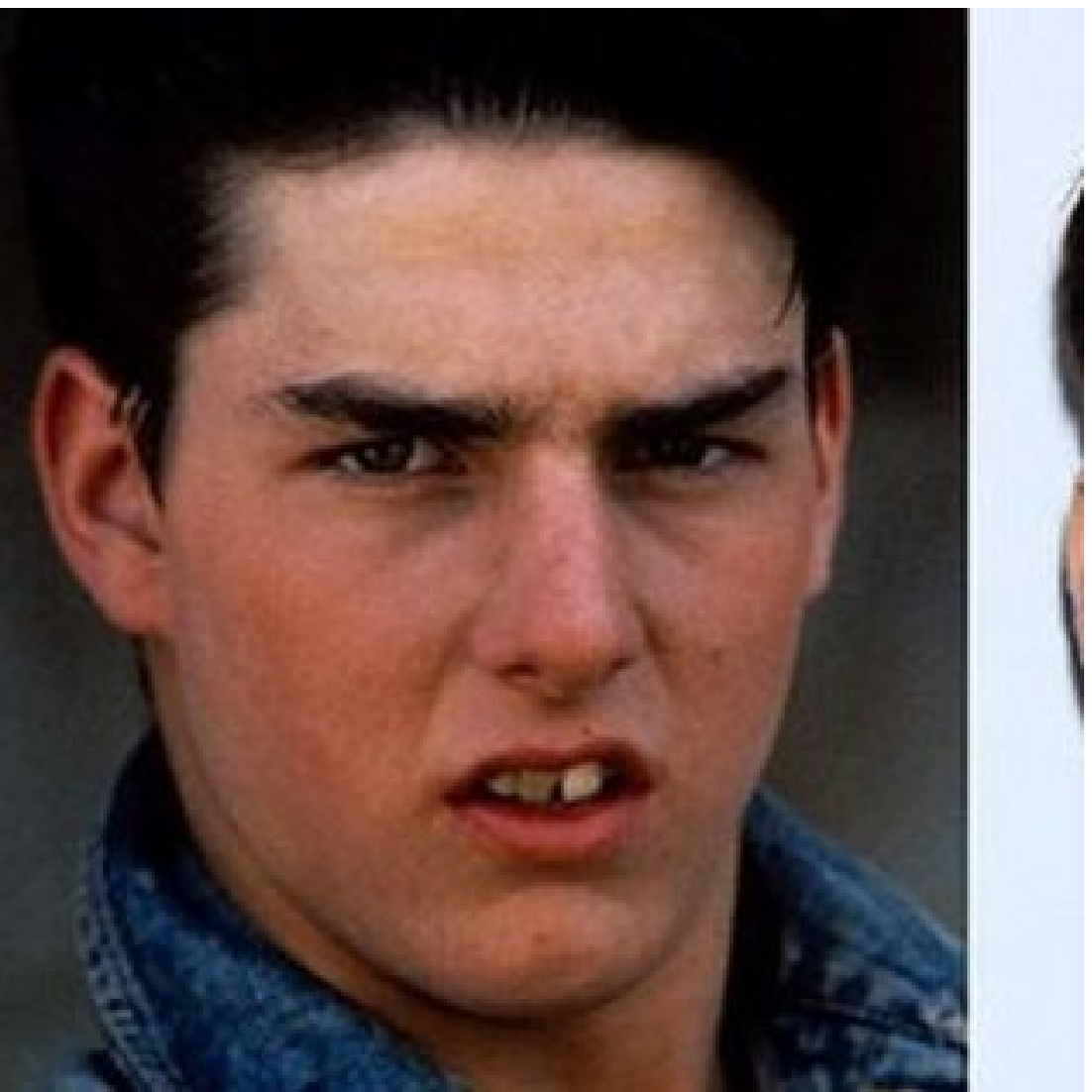}\label{fig:WA3}}
  \subfigure[Anger $\rightarrow$ Disgust]{\includegraphics[width=13mm]{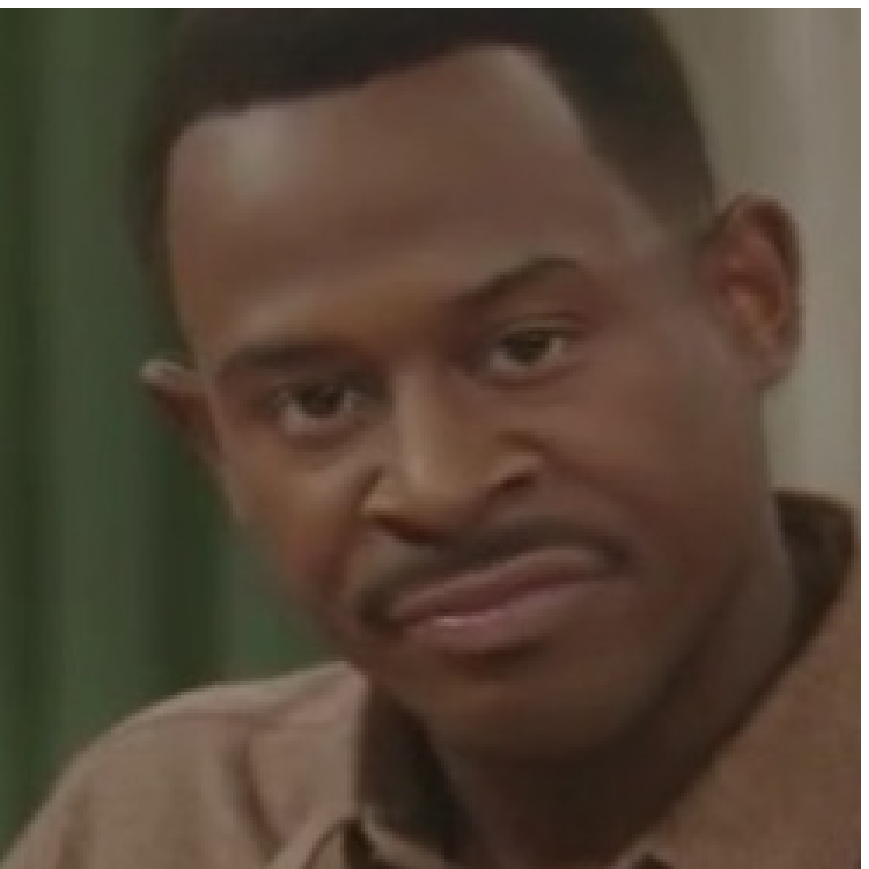}\label{fig:WA4}}
  \subfigure[Anger $\rightarrow$ Disgust]{\includegraphics[width=13mm]{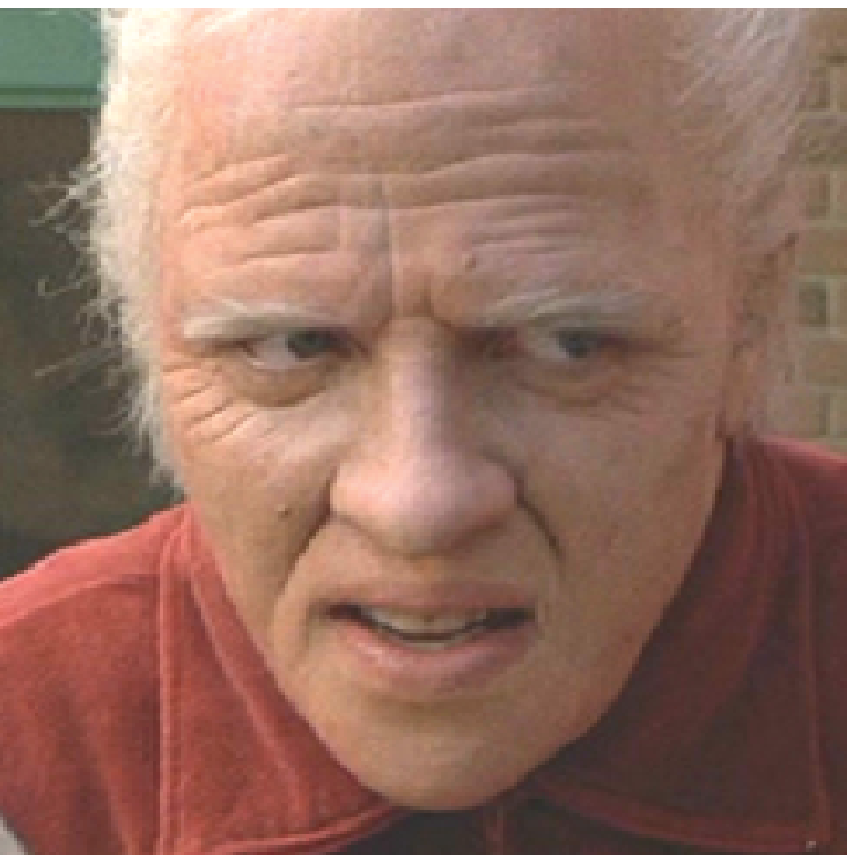}\label{fig:WA5}}
  \subfigure[Anger $\rightarrow$ Disgust]{\includegraphics[width=13mm]{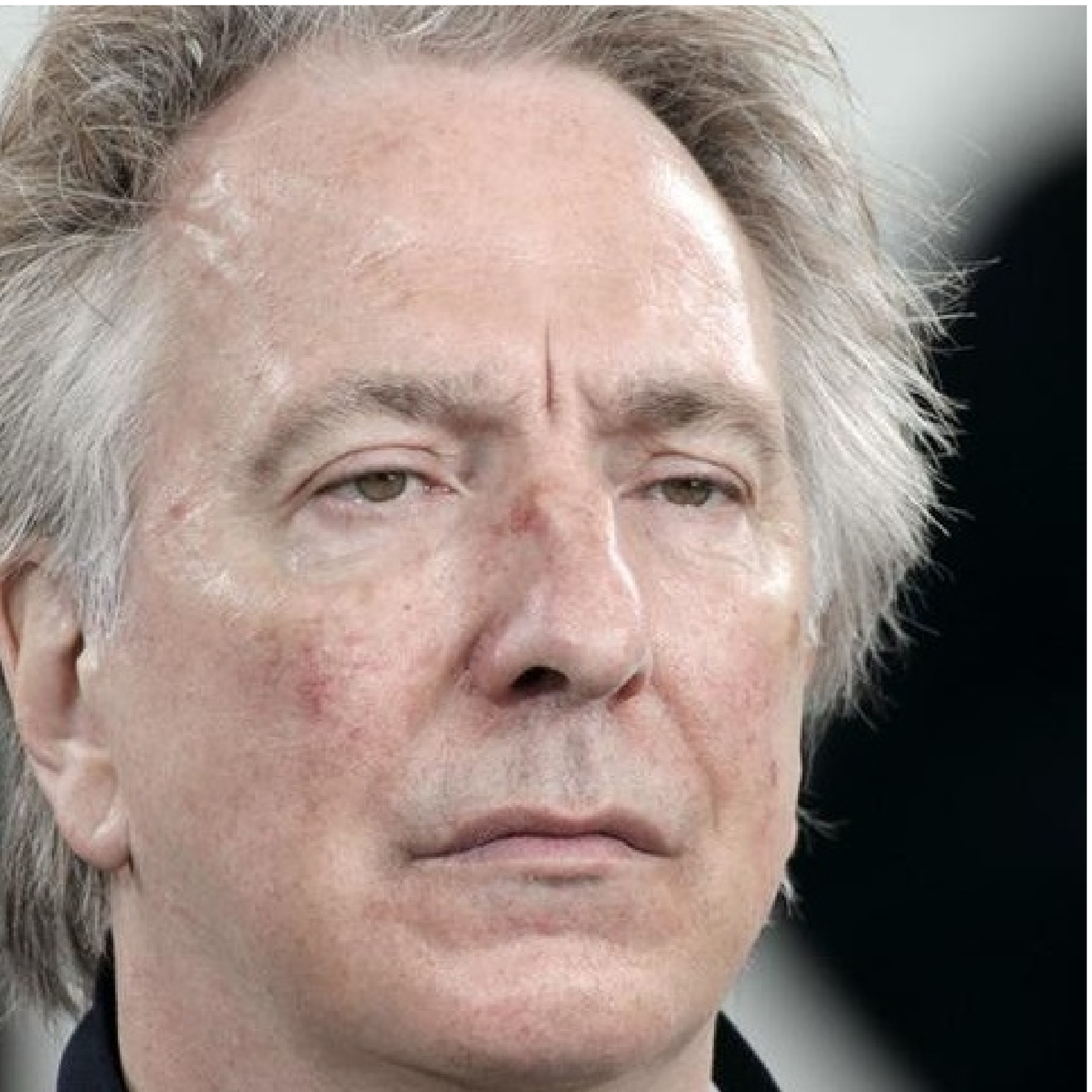}\label{fig:WA6}}
  \caption{Examples of mis-classified categories}
  \label{fig:sample_wrong}
\vspace{-5mm}
\end{figure}

Table~\ref{tab:classification_ratio_cf} shows the confusion matrix for the classification results in Table~\ref{tab:classification_ratio}. The element in confusion matrix indicates the matched number of the predicted category and the labeled category for eight emotion faces. Fig.~\ref{fig:sample_wrong} shows some examples of the mis-classified cases. From Table~\ref{tab:classification_ratio_cf}, we found that many mis-classified cases where `Anger' was mis-classified to `Disgust'.

\begin{table}[tbp]
  \caption{Classification accuracy for AffectNet}
\vspace{-3mm}
\label{tab:classification_ratio}
\begin{center}
\scalebox{0.9}[0.9]{
\begin{tabular}{l|r|r|r}
\hline \hline
& \multicolumn{1}{c|}{CNN(\cite{AffectNet})} & \multicolumn{2}{c}{Adaptive DBN} \\ \cline{2-4}
\multicolumn{1}{c|}{Category} & Test data & Train data & Test data \\ \hline
\hline
Neutral   & 63.0\% & 99.3\% &  87.8\% \\ \hline
Happy     & 88.0\% & 99.9\% &  92.4\% \\ \hline
Sad       & 63.0\% & 99.2\% &  84.2\% \\ \hline
Surprise  & 61.0\% & 99.4\% &  85.8\% \\ \hline
Fear      & 52.0\% & 99.5\% &  90.4\% \\ \hline
Disgust   & 52.0\% & 99.3\% &  92.4\% \\ \hline
Anger     & 65.0\% & 98.2\% &  78.4\% \\ \hline
Contempt  &  8.0\% & 98.8\% &  87.6\% \\ \hline
\hline
\end{tabular}
} 
\end{center}
\vspace{-5mm}
\end{table}

\begin{table*}[tbp]
  \caption{The confusion matrix for classification results by Adaptive DBN (Num.)}
\vspace{-3mm}
\label{tab:classification_ratio_cf}
\begin{center}
\scalebox{0.8}[0.8]{
\begin{tabular}{l|l|r|r|r|r|r|r|r|r}
\hline
\multicolumn{2}{c|}{} & \multicolumn{8}{c}{Predicted Category} \\ \cline{3-10}
\multicolumn{2}{c|}{} & Neutral & Happy & Sad & Surprise & Fear & Disgust & Anger & Contempt \\ \hline 
\multirow{8}{*}{\rotatebox[origin=c]{90}{True Category}}& Neutral & 439 & 2 & 7 & 5 & 8 & 16 & 4 & 19 \\
& Happy & 7 & 462 & 2 & 0 & 4 & 12 & 1 & 12 \\
& Sad & 12 & 3 & 421 & 13 & 11 & 20 & 5 & 15 \\
& Surprise & 15 & 4 & 10 & 429 & 11 & 22 & 0 & 9 \\
& Fear & 10 & 2 & 10 & 10 & 452 & 8 & 3 & 5 \\
& Disgust & 8 & 2 & 3 & 5 & 8 & 462 & 5 & 7 \\
& Anger & 14 & 4 & 8 & 10 & 9 & 47 & 392 & 16 \\
& Contempt & 17 & 8 & 6 & 3 & 2 & 21 & 5 & 438 \\
\hline 
\end{tabular}
} 
\end{center}
\vspace{-5mm}
\end{table*}

\begin{table}[tbp]
\caption{Training data for child model\cite{Ichimura19}}
\vspace{-3mm}
\label{tab:relearning_dataset}
\begin{center}
\scalebox{0.9}[0.9]{
\begin{tabular}{c|l|r}
\hline 
data set & \multicolumn{1}{c|}{Description} & number  \\ 
\hline
Set 0 & All images for Anger and Disgust &   1,000 \\ 
Set 1 & The images correctly classified at Set 0 &  854 \\ 
Set 2 & The images incorrectly classified Set 0 &  146 \\ 
\hline
\end{tabular}
} 
\end{center}
\vspace{-5mm}
\end{table}

\begin{table}[tbp]
\caption{KL Divergence\cite{Ichimura19}}
\vspace{-4mm}
\label{tab:kl}
\begin{center}
\begin{tabular}{l|r}
\hline 
\multicolumn{1}{c|}{Model} & KL \\ \hline
$D_{KL}(P, Q1)$   & 0.188 \\
$D_{KL}(P, Q2)$   & 0.660 \\ 
\hline
\end{tabular}
\end{center}
\vspace{-5mm}
\end{table}

In \cite{AffectNet}, the matching ratio of the labeled data by two annotators to an image for each category was reported. Table~\ref{tab:ref} shows ratio that two annotators' answer for an image is same. From Table~\ref{tab:ref}, the ratio is not surely high, with the highest being `Happy' at 79.6\%, while that of `Neutral,' `Contempt,' and `Anger' is low at 50.8\%, 66.9\% and 62.3\%. In addition, the agreement degree is moderate according to the Cohen's kappa coefficient \cite{Cohen60}, calculated from Table~\ref{tab:ref}, but the data set contains subjectivity of multiple annotators.

For the ensemble learning of two or more DBNs, we constructed a parent DBN which was trained for the given data set and two or more child DBNs which were trained for the mis-classified test cases. The KL divergence with the original parent model $T$ and the child model $S$ was calculated by the following equations.

\vspace{-2mm}
\begin{equation}
\label{eq:kl_f}
D_{KL}(P_{T} || P_{S}) = \sum_{i} P_{T}(x_{i}) \log P_{T}(x_{i})/P_{S}(x_{i}),
\vspace{-1mm}
\end{equation}
\begin{equation}
\label{eq:kl_b}
D_{KL}(P_{S} || P_{T}) = \sum_{i} P_{S}(x_{i}) \log P_{S}(x_{i})/P_{T}(x_{i}),
\vspace{-2mm}
\end{equation}
where $P_{T}(x_{i})$ and $P_{S}(x_{i})$ are the output probability distribution for the input $x_{i}$ at the model $T$ and $S$, respectively.

In \cite{Ichimura19}, we constructed a child model for mis-classified data and correctly labeled data for `Anger' and `Disgust'. As shown in Table \ref{tab:relearning_dataset}, for `Anger' and `Disgust', we define $Set 0$ is the all images, $Set 1$ is the correctly classified cases, and $Set 2$ is the incorrectly classified data. The parent model $P$, and the child models $Q1, Q2$ were trained for the data set $Set 0, Set 1, Set2$, respectively.

Table~\ref{tab:kl} shows the KL divergence of parent model $P$ and child models $Q1$ and $Q2$. KL divergence of $P$ and $Q2$ was larger than that of $P$ and $Q1$. The result means a certain difference between the models. KL divergence for each sample of $P$ and $Q2$ was measured and its range was from $0.0000$ to $0.0025$. Therefore, we assumed there are the three levels of KL divergence in the data and defined three threshold values $\theta_{KL} = \{0.0010, 0.0015, 0.0020\}$ as the boundary of two emotion types for the mis-classified data. The three child models were trained for the mis-classified data divided by the three threshold values, respectively. As a result, the classification accuracy were 95.8\%, 97.2\%, and 95.2\% for the three threshold values. The child model with the highest accuracy was trained for data with $\theta_{KL} = 0.0015$.

We consider a generation method of new neurons at the parent DBN model by using the threshold value of KL divergence so that the annotator's mis-classified data is divided into two subsets. The improvement of the network structure at the parent model will reduce the calculation time of ensemble learning of deep learning.

\subsection{KL divergence based Distillation learning model}
\label{sec:Remodeling}
\begin{figure}[tbp]
  \vspace{-5mm}
  \centering
  \includegraphics[scale=0.39]{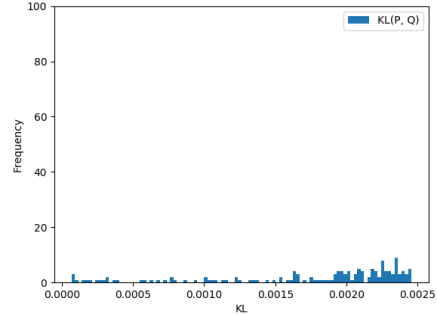}
  \vspace{-5mm}
  \caption{The histogram of KL(P, Q)}
  \label{fig:kl-p-q}
  \vspace{-5mm}
\end{figure}

In this paper, we define a child model $Q$, which initializes the network structure to copy weights from the parent $P$. The child model $Q$ trains the cases for $P$'s mis-classified `Anger' and `Disgust'. As a result of the re-learning, the classification ratio of the child model $Q$ will be improved perfectly. The KL divergence of $P$ and $Q$ for an input $x_{i}$ was calculated by Eq.~(\ref{eq:kl_PQ}).
\begin{equation}
\label{eq:kl_PQ}
D_{KL}(P, Q) = \sum_{i} P(x_{i}) \log P(x_{i})/Q(x_{i}),
\vspace{-2mm}
\end{equation}
The KL divergence between $P$ and $Q$ was $0.660$. Fig.~\ref{fig:kl-p-q} shows the KL divergence of two models for the cases of `Anger' and `Disgust'. From Fig.~\ref{fig:kl-p-q}, KL divergence for the mis-classified data was varied within the range of threshold value $0.000$ to $0.0025$ in this experiment.

\begin{figure}
  \centering
  \subfigure[Parent model]{\includegraphics[scale=0.45]{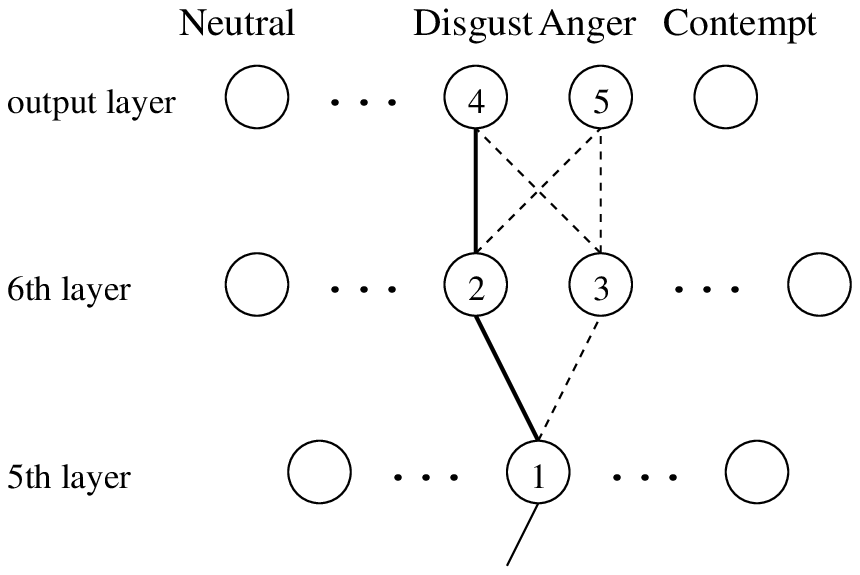}\label{fig:path_p}}
  \subfigure[Child model]{\includegraphics[scale=0.45]{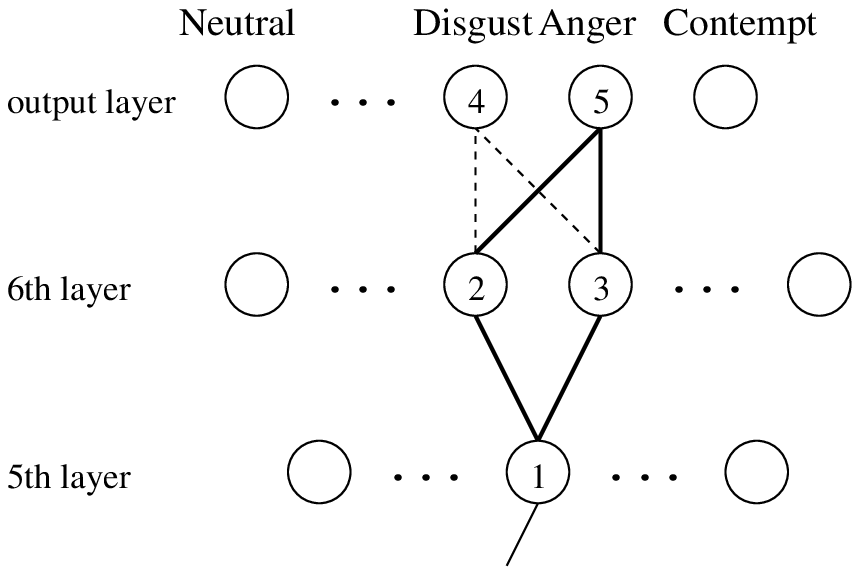}\label{fig:path_q}}
  \caption{Firing path in the DBN}
  \label{fig:result_path_diff}
\end{figure}

\begin{algorithm}
\caption{Fine Tuning Method}
\label{alg:finetuning-org}                        
\begin{algorithmic}[1]
\STATE The $L$ layers network structure by training Adaptive DBN was constructed.
\STATE Fine tuning method described in {\bf Algorithm \ref{alg:finetuning-l}} at the $l$-th layer ($l \leq L$)was implemented from the first layer to the $L$-th layer.
\end{algorithmic}
\end{algorithm}
\begin{algorithm}[tb]
\caption{Procedure of modifying weight at the $l$-th layer}
\label{alg:finetuning-l}                        
\begin{algorithmic}[1]
 \STATE Let $X(\bvec{x}_{1}, \cdots, \bvec{x}_p ,\cdots, \bvec{x}_{N})$ be input patterns, where $N$ is the number of patterns. Let $Y(y_{1}, \cdots, y_p ,\cdots, y_{N})$ be teacher signals. The trained DBN has $L$ layers. Let $l$ is the target layer for fine tuning.
\STATE Give $X$ to the trained network for feed forward calculation. In the calculation, save information which neuron was activated from lower to upper layer. Let $X^{T}$ and $X^{F}$ be correct and wrong input patterns for $Y$, respectively.
\STATE In the $l$-th layer, if a neuron $j$ is satisfied with Eq.~(\ref{eq:condition_A}), modify weights connected to the neuron $j$ to $w^{correct}$. Eq.~(\ref{eq:condition_A}) is the ratio that only $X^{T}$ is fired at the neuron $j$. $w^{correct}$ is a constant value (e.g. $w^{correct} = 1$).
\begin{equation}
\label{eq:condition_A}
 |Act^{T}_{j}|/(|X^{T}| + |X^{F}|) \geq \theta^{T},
\end{equation}
where, $\theta^{T}$ is the threshold value. In this paper, we assume that $\theta^{T}$ is $0.3$ from the empirical studies.
\STATE In the $l$-th layer, if a neuron $j$ is satisfied with Eq.~(\ref{eq:condition_B}), modify weights connected to the neuron $j$ to $w^{wrong}$. Eq.~(\ref{eq:condition_B}) is the ratio that only $X^{F}$ is fired at the neuron $j$. $w^{wrong}$ is a constant value (e.g. $w^{wrong} = 0$).
\begin{equation}
\label{eq:condition_B}
|Act^{F}_{j}|/(|X^{T}| + |X^{F}|) \geq \theta^{F},
\end{equation}
where, $\theta^{F}$ is the threshold value. In this paper, we assume that $\theta^{T}$ is $0.3$ from the empirical studies.
\end{algorithmic}
\end{algorithm}

In addition to the difference of KL divergence, the difference of network signal flow for given input pattern was also investigated. The paths passed from input layer to output layer were explored. If the wrong output was calculated for the test case, the neuron in the path was activated incorrectly. The method is utilizing that a hidden neuron of RBM is represented as binary patterns \{0, 1\}. All possible combinations of binary pattern \{0, 1\} are given to the trained DBN and the signal flow of the network from the lower layer to the upper layer is analyzed to find the activated path. The fine tuning method can repair the activated path of network partially for improving the classification accuracy. {\bf Algorithm \ref{alg:finetuning-org}-\ref{alg:extract_rule}} show the procedures of the fine tuning and knowledge extraction method. The method can discover the different paths for the parent model $P$ and the child model $Q$. (See details in \cite{Kamada17_IJCIStudies, Kamada19_WorldScientific}).

\begin{algorithm}[tbp]
\caption{Knowledge Extraction by C4.5\cite{Quinlan96}}
\label{alg:extract_rule}                        
\begin{algorithmic}[1]
\STATE Let $X(\bvec{x}_{1}, \cdots, \bvec{x}_p ,\cdots, \bvec{x}_{N})$ be input patterns, where $N$ is the number of patterns. $\bvec{x}_{p}$ is a $I$-dimensional vector. The trained DBN has $L$ layers. Let $Net()$ be the function which calculates output patterns $Y(y_{1}, \cdots, y_p, \cdots, y_{N})$ for given $X$. $y_{p}$ has a class value for classification.
\STATE Give $X$ to $Net()$ and calculate $Y$.
\STATE Extract rules by C4.5.
\end{algorithmic}
\end{algorithm}

As a result, there was no significant difference in the path in the lower layer closer to the input layer, but there were characteristic differences in the upper layer near the output layer as shown in Fig.~\ref{fig:result_path_diff}. Fig.~\ref{fig:path_p} and Fig.~\ref{fig:path_q} show the paths from the fifth layer to the output layer of the parent model and the child model, respectively. In Fig.~\ref{fig:result_path_diff}, a node indicates a neuron, a line between nodes indicates a weight, the bold line indicates the path that actually activated, and the dotted line indicates the path that does not activated. Fig.~\ref{fig:path_p} shows an example that the input signal passed through 1, 2, and 4 neurons from the fifth layer to the output layer and then mis-classified to `Disgust' in the parent model $P$. On the other hand, in Fig.~\ref{fig:path_q}, the child model $Q$ also followed the same path as the parent model, but in addition to the path of the parent model, the path from 1 to 3 neurons also activated which was not activated in the parent model. Due to the combination of these activated patterns, `Anger' was improved to output instead of `Disgust' in the child model $Q$.

The difference of signal flows among two models will occur due to the discrepancy of representing power of features between the parent model and the child model. If the activated neuron path in the child model as shown in Fig.~\ref{fig:kl-p-q} is also generated by the neuron generation algorithm, the parent model itself can represent the different pattern as same as the child model. Based on the observation, we found that such different paths are seen with large KL divergence. In this paper, we construct the parent model by copying the corresponding neuron in the child model with same weights between neurons, if the larger KL divergence both the parent model and the child model than the certain value $\theta_{KL}$ and the different path between two models is discovered as shown in Fig.~\ref{fig:result_path_diff}. The parent model is trained with small perturbation after the new neuron is inserted at the corresponding position. Of course, it is difficult to find the appropriate $\theta_{KL}$ automatically, but we can determine it by computing KL divergence and the number of mis-classified cases in near future. 

Table~\ref{tab:classification_ratio_relearning} shows the classification results of re-modeling for the test data set. In this paper, we set $\theta_{KL} = 0.0015$ based on the experimental result in \cite{Ichimura19}. Before re-modeling, the correct classification ratios for `Disgust' and `Anger' were 92.4\% and 78.4\%, respectively, but those ratios improved to 94.7\% and 91.3\% by re-modeling. Table~\ref{tab:classification_ratio_relearning} shows only the relevant results of `Disgust' and `Anger', but the classification accuracy of other categories did not change in the experiment because of the $\theta_{KL}$ based on the results of preliminary experiments. In addition, we note the computational time of training the model was 20.32 hours, while that of the reconstruction using child models and KL divergence was 2.95 hours, in our environment using two RTX 2080Ti GPUs.

\begin{table}[tbp]
\caption{The classification rate of the distillation model}
\vspace{-3mm}
\label{tab:classification_ratio_relearning}
\begin{center}
\begin{tabular}{l|r|r}
\hline \hline
\multicolumn{1}{c|}{Category} & Before & After \\ \hline
\hline
Disgust   & 92.4\% & 94.7\% \\ \hline
Anger     & 78.4\% & 91.3\% \\ \hline
\hline
\end{tabular}
\end{center}
\vspace{-5mm}
\end{table}

\section{Conclusive Discussion}
\label{sec:ConclusiveDiscussion}
In this paper, we employ Adaptive DBN that finds the optimal structure by generating/annihilating neurons and generating layers during learning to the AffectNet: Facial Expression Image database. The classification results for training data set was perfectly, however, that for certain categories in test data set cannot reach high ratio, because human emotion contains many complex and ambiguous features due to the annotated labels depending on human subjectively. In order to improve them, the parent model and the child model were constructed, and then KL divergence between two models was measured. A re-modeling method was proposed to make fine tuning the trained DBN, and then the re-trained model with the new neurons was able to improve classification accuracy for 'Anger' and 'Disgust'. Beside these cases, there are some mis-classified cases (e.g. 'Contempt'), because they contain ambiguous human emotions as same as 'Anger' and 'Disgust'. We will investigate them in future work. 

\section*{Acknowledgment}
This work was supported by JSPS KAKENHI Grant Number 19K12142, 19K24365, and obtained from the commissioned research by National Institute of Information and Communications Technology (NICT, 21405), JAPAN.


\begin{thebibliography}{00}
\bibitem{Bengio09}
Y.Bengio, \emph{Learning Deep Architectures for AI}, Foundations and Trends in Machine Learning archive, vol.2, no.1, pp.1-127 (2009).

\bibitem{Quoc12}
V.Le.Quoc, R.Marc's Aurelio, et.al., \emph{Building high-level features using large scale unsupervised learning}, Proc. of 2013 IEEE International Conference on Acoustics, Speech and Signal Processing, pp.8595-8598 (2013).

\bibitem{AlexNet}
 A.Krizhevsky, I.Sutskever, G.E.Hinton, \emph{ImageNet Classification with Deep Convolutional Neural Networks}, Proc. of Advances in Neural Information Processing Systems 25 (NIPS 2012) (2012).

\bibitem{GoogLeNet15}
C.Szegedy, W.Liu, et.al., \emph{Going Deeper with Convolutions}, 2015 IEEE Conference on Computer Vision and Pattern Recognition (CVPR), pp.1-9 (2015).

\bibitem{VGG16}
K.Simonyan, A.Zisserman, \emph{Very deep convolutional networks for large-scale image recognition}, Proc. of International Conference on Learning Representations (ICLR 2015) (2015).
  
\bibitem{ResNet}
K.He, X.Zhang, S.R en, J.Sun, \emph{Deep residual learning for image recognition}, Proc. of 2016 IEEE Conference on Computer Vision and Pattern Recognition (CVPR), pp.770-778 (2016).

\bibitem{Hinton06}
G.E.Hinton, S.Osindero and Y.Teh, \emph{A fast learning algorithm for deep belief nets}, Neural Computation, vol.18, no.7, pp.1527-1554 (2006).

\bibitem{Hinton12}
G.E.Hinton, \emph{A Practical Guide to Training Restricted Boltzmann Machines}, Neural Networks, Tricks of the Trade, Lecture Notes in Computer Science (LNCS, vol.7700), pp.599-619 (2012).

\bibitem{Kamada16_SMC}
S.Kamada and T.Ichimura, \emph{An Adaptive Learning Method of Restricted Boltzmann Machine by Neuron Generation and Annihilation Algorithm}, Proc. of 2016 IEEE International Conference on Systems, Man, and Cybernetics (IEEE SMC 2016), pp.1273-1278 (2016).

\bibitem{Kamada16_ICONIP}
S.Kamada and T.Ichimura, \emph{A Structural Learning Method of Restricted Boltzmann Machine by Neuron Generation and Annihilation Algorithm}, Neural Information Processing, vol.9950 of the series Lecture Notes in Computer Science, pp.372-380 (2016).

\bibitem{Kamada16_TENCON}
S.Kamada and T.Ichimura,\emph{An Adaptive Learning Method of Deep Belief Network by Layer Generation Algorithm}, Proc. of IEEE TENCON2016, 2971/2974 (2016).


\bibitem{CIFAR10}
A.Krizhevsky, \emph{Learning Multiple Layers of Features from Tiny Images}, Master of thesis, University of Toronto (2009).


\bibitem{Kamada18_Springer}
S.Kamada, T.Ichimura, A.Hara, and K.J.Mackin, \emph{Adaptive Structure Learning Method of Deep Belief Network using Neuron Generation-Annihilation and Layer Generation}, Neural Computing and Applications, pp.1-15 (2018).

\bibitem{Kamada19_WorldScientific}
S.Kamada, T.Ichimura, T.Harada, \emph{Knowledge Extraction of Adaptive Structural Learning of Deep Belief Network for Medical Examination Data}, International Journal of Semantic Computing, Vol.13, No.1, pp. 67-86 (2019).

\bibitem{AffectNet}
A.Mollahosseini, B.Hasani, M.H.Mahoor: \emph{AffectNet: A Database for Facial Expression, Valence, and Arousal Computing in the Wild}, IEEE Transactions on Affective Computing, vol.10, No.1 pp.18-31 (2017).

\bibitem{Ichimura19}
T.Ichimura, S.Kamada, \emph{Re-learning of Child Model for Misclassified data by using KL Divergence in AffectNet: A Database for Facial Expression}, Proc. of 2019 IEEE 11th International Workshop on Computational Intelligence and Applications (IWCIA2019), pp.15-20 (2019).

\bibitem{Cohen60}
J.Cohen, \emph{A coefficient of agreement for nominal scales}, Educational and Psychological Measurement, vol.20, no.1, p.37--46, (1960).
  
\bibitem{Kamada17_IJCIStudies}
S.Kamada and T.Ichimura, \emph{Fine Tuning of Adaptive Learning of Deep Belief Network for Misclassification and its Knowledge Acquisition}, International Journal Computational Intelligence Studies, Vol.6, No.4, pp.333--348 (2017).

\bibitem{Quinlan96}
J.R.Quinlan, \emph{Improved use of continuous attributes in c4.5}, Journal of Artificial Intelligence Research, No.4, pp.77-90 (2016).
\end{thebibliography}
\end{document}